# Integration of Rule Based Expert Systems and Case Based Reasoning in an Acute Bacterial Meningitis Clinical Decision Support System

Mariana Maceiras Cabrera

Departamento de Informática y Ciencias de la Computación
Universidad Católica del Uruguay
Montevideo, Uruguay

Ernesto Ocampo Edye

Departamento de Informática y Ciencias de la Computación
Universidad Católica del Uruguay
Montevideo, Uruguay

*Abstract*—**This article presents the results of the research carried out on the development of a medical diagnostic system applied to the Acute Bacterial Meningitis, using the Case Based Reasoning methodology. The research was focused on the implementation of the adaptation stage, from the integration of Case Based Reasoning and Rule Based Expert Systems. In this adaptation stage we use a higher level RBC that stores and allows reutilizing change experiences, combined with a classic rule-based inference engine. In order to take into account the most evident clinical situation, a pre-diagnosis stage is implemented using a rule engine that, given an evident situation, emits the corresponding diagnosis and avoids the complete process.**

*Keywords-Case-Based Reasoning; Medical Diagnosis; Expert Systems; Acute Bacterial Meningitis*

## I. INTRODUCTION

Clinical Decision Support Systems (hereafter CDSSs) have become paramount in the medical environment, providing support to the clinical expert in the diagnosing process, mainly assisting the analysis and synthesis of bulk information in little time. These systems enable the reduction of the intrinsic degree of uncertainty that any decision-making process in the medical environment entails. [1], [2]

Several aspects related to a specific kind of CDSS, Medical Diagnostic Systems (hereafter MDSs), will be addressed throughout this document. From a set of symptoms and signs shown by the patient, the MDS are capable of **suggesting** a set of differential diagnoses, possibly with a certainty degree associated to each of them. [1], [3]

While developing MDSs, either in their construction and application, several problems arise: 1) Representation of the domain knowledge; 2) Processing of such knowledge; 3) Obtaining results that are accurate enough and 4) Presentation of the results to the user, and their explanation. Each of these problems is solved in a particular way, depending on the method applied to the development of the system. In this document is presented an alternative way to solve these problems using Case Based Reasoning. [1], [4]

This research is focused on the development of an Acute Bacterial Meningitis MDS integrating Case Based Reasoning (hereafter CBR) with Rule Based Expert Systems (hereafter RBR).

This document is organized in the following way: in Section 2 is presented the case study used in the research: Diagnosis of Acute Bacterial Meningitis, followed in section 3 by the presentation of the addressed problem. Section 4 presents a brief summary of state-of-the-art techniques and referred applications. Afterwards, section 5 describes the proposed computer solution in detail, while in section 6 is described the composition of the knowledge base, which represents reality and is used in simulations, and the way it was built. Section 7 presents the results of the system's performance assessment and section 8 outlines the conclusions of this work.

## II. CASE STUDY: DIAGNOSIS OF ACUTE BACTERIAL MENINGITIS

The medical diagnosing task consists in "translating" the relation between a set of signs and symptoms found in a patient, and the possible pathologies she or he might suffer from. [3], [1] This work addresses the medical diagnosis of the disease known as Acute Bacterial Meningitis (hereafter ABM).

ABM is an infectious disease caused by bacteria that attack the meninges (a membrane system that covers the Central Nervous System – brain and spinal cord). The meningitis is an inflammatory process of the leptomeninges (pia mater and arachnoid mater), and the cerebrospinal fluid (CSF) contained inside them [5]. The Meningitis poses a great threat to optical, facial and auditory nerves, and other kinds of neurologic sequels.

In this research we have taken as a reference a subset of the typical signs and symptoms of this disease – a total of 81 symptoms have been considered - which are: convulsions, depression, fever, hypertense fontanelle, nape stiffness, trunk stiffness, skin purpuric syndrome, vomits, somnolence, irritability, facial paralysis, cervical adenopathies, haemocultivation with bacteria, bacteria in CSF, muscular hypotonicity, cloudy aspect CSF, clear aspect CSF, hydrocephaly in ecography, tumors in tomography, among many others. [6], [7].

The combination of the presence of several of these symptoms allows identifying the disease under study, but it could also be indicative, to a greater or lesser extent, of other diseases identified as "differential diagnoses". The relevant differential diagnoses of the disease of ABM are: Acute Viral Meningitis, Tuberculous Meningitis, Encephalitis, Brain Abscess, Meningism, Meningeal reaction to nearby inflammation, Meningeal Haemorrhage and Brain Tumor. [7]. The task of the doctor is to diagnose accurately the disease within these alternatives.

### III.   PROBLEM OUTLINE

This research focuses on the development of an MDS for the disease known as ABM, integrating the CBR and RBR methods, making special emphasis on the implementation of the adaptation of the CBR cycle stage. This stage is fundamental in a case-based MDS, given that once recovered the most similar case, it is highly probable to find differences in the descriptions of the problems. This could then indicate differences in the solutions, which entails a possible error in the diagnosis.

On the other hand, the adaptation capabilities of a CBR system applied to a CDSS allow its usage by not so experienced physicians. This fosters its utility as a support tool also in the medical learning and practice.

### IV.   THEORETICAL FRAMEWORK

#### A. Clinical Decision Support Systems

A Clinical Decision Support System (CDSS) is an expert system that provides support to certain reasoning tasks, in the context of a clinical decision. [1]

A medical diagnostic decision support system – MDS - is defined as a computer algorithm aimed at assisting the doctor in one or more stages that comprise the medical diagnostic process. [1]

One of the first CDSS that appeared in the marketplace is MYCIN, a system developed at Stanford University. This system was designed to diagnose and recommend treatments for blood infections. [8], [9]

Other systems of interest are: IMAGECREEK – image analysis - [10], CADI – medicine students' tutorial -  [11], SCINA – diagnosis of cardiac diseases - [12], CARE-PARTNER – diagnosis and treatment scheduling of stem cells transplant - [13], AUGUSTE – diagnosis and scheduling of Alzheimer treatment - [14], T-IDDM – treatment scheduling of diabetes - [15].

There are many computer techniques and methods – especially in the Artificial Intelligence field – that have been applied in the last 30 years or more in the development of systems of this nature: pattern analysis, neural networks, expert systems and Bayesian networks, among others.

The rule based reasoning is one of the most used techniques [16], [17], and in recent years case-based reasoning has gained much importance in this field [18], [19], [20], [4].

#### B. Case-Based Reasoning (CBR)

Case Based Reasoning is a methodology for problem solving that focuses on the reutilization of past experience. It is based on solutions, information and knowledge available in similar problems previously solved. The implementation of this method requires the existence of a knowledge base that contains the cases that contain previous experience. It is also necessary to count with a mechanism that allows us to infer, for a new case, the solution based on foregoing cases. CBR's basic principle is that *similar problems* will have *similar solutions.* [19], [20], [21], [4], [18]

In CBR the structures handled are known as *cases.* A case represents a problem situation. It could be more formally defined as contextualized knowledge that represents past experience, and implies an important teaching to accomplish the objectives of a rational agent. [19], [20], [18], [22]

In Figure 1 is presented the CBR cycle. It can be observed the presence of 4 sub-processes or stages ("Retrieve", "Reuse", "Revise" and "Retain") that explicate the operation of this methodology.

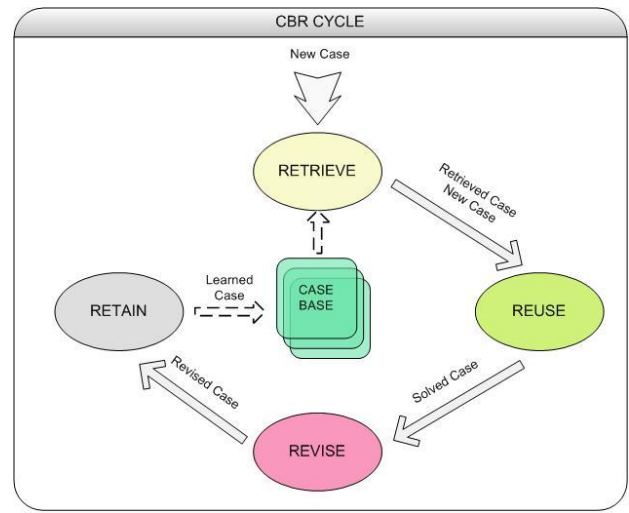

Figure 1.   CBR Cycle

First, when a new problem is posed to the system, it proceeds to identify correctly the current case, and *retrieve* the most *similar* cases from the knowledge base.

Once the cases have been obtained, the information and knowledge contained in them is reused to solve the new problem, from which a solution to propose to the user is reached. This stage might possibly entail an *adaptation* process of the solution of the retrieved case, considering the differences between both problematic situations.

Also, these systems include a stage of proposed *solution assessment,* and a later *learning* from this case. Then, after the *reusage* stage it proceeds to the proposed solution *revision* process, which is assessed and *repaired* if it fails, obtaining a *confirmed or revised solution.*

After that, the ***learning*** stage takes place. In this stage the information and knowledge obtained with the solution of the new case is ***retained***, so as to be useful to solve future problems. This stage implicates a modification of the knowledge base, or of the methods to retrieve and reuse cases from such base.

These systems progressively enlarge their knowledge bases with new cases while new problems are solved, gradually widening the range of situations in which the system is applicable and the accuracy of the proposed solutions. Once finalized the retention the cycle can start all over again. [19], [20]

Regarding the ***reutilization or adaptation*** methods – main objectives of this research -, it is necessary to consider [19]:

- Determine the differences between the new case and the previous case recovered.

- Determine the aspects of previous cases that can be reutilized without changes.

- Determine how to adapt a solution of a previous case based on the differences observed with the new case.

The adaptation arises from: 1) the description of the new problem, 2) the solution provided for a similar problem, and optionally 3) the description of the corresponding similar problem. [18]

The adaptation process must take into account two key issues: 1) How to identify what has to be adapted (identifying the adaptation need) and 2) how to choose an appropriate method to carry out a required adaptation. This last issue also implicates: identifying what must be changed, finding an appropriate adaptation technique or strategy, and choosing among a set of adequate strategies. [18], [4].

The most relevant adaptation techniques can be classified in three main groups: **a) Substitution methods:** null adaptation, re-instantiation, parameter adjustment, adjustment based on a fuzzy decision structure, local search, query memory, specialized search, case based substitution, restriction based substitution, user feedback based substitution; **b) Transformation methods:** common-sense transformation, model guided transformation and **c) Other methods**: adaptation and repairment with specific purposes, derivational replay, general framework for adaptations through substitution, case based adaptation, hierarchy based adaptation, compositional adaptation. [18], [20], [23], [22], [21], [24], [4], [25].

Each of these techniques or methods differ in complexity taking into account two issues: what is to be changed in the previous solution, and how will that change be achieved [21], [26]

### C. CBR Systems Examples

- DIAL [27], [28], [29], [30], [31], is a planning system for the recovery from disasters that stands out for its adaptation stage, that uses a combination of CBR and rules. It also implements the adaptation guided retrieval.

- Deja Vu [32], is a system dedicated to the automatic programming of machinery in a control plant. In this system, its adaptation guided retrieval approach is fundamental.

### D. Rule Based Expert Systems (RBR)

An RBR system has an inference engine that uses rules to reach conclusions based on premises and a certain context state. These systems are comprised of three main parts: inference engine, rule base, and working memory (auxiliary space in the memory to be used during the reasoning). [16], [17]

An example of a rule based CDSS is MYCIN, one of the first systems to appear in the marketplace. This system counts with a knowledge base composed of IF-THEN rules, which are associated a certainty factor. Through these factors it can be given each rule different weights so as to determine their influence. [8], [9]

Another example of a rule based CDSS is CASEY, which is an hybrid system that combines CBR and RBR. CASEY is a medical diagnostic system that offers a causal explanation of the patient's pathology. It is applied to heart failures diagnose. This system outstands by its use of rules, in both retrieving and reusing steps. [33]

## V. PROPOSED SOLUTION

A hybrid system has been developed, that combines CBR and RBR methodologies: SEDMAS-RBC-Adapt, whose operation is depicted in Figure 2.

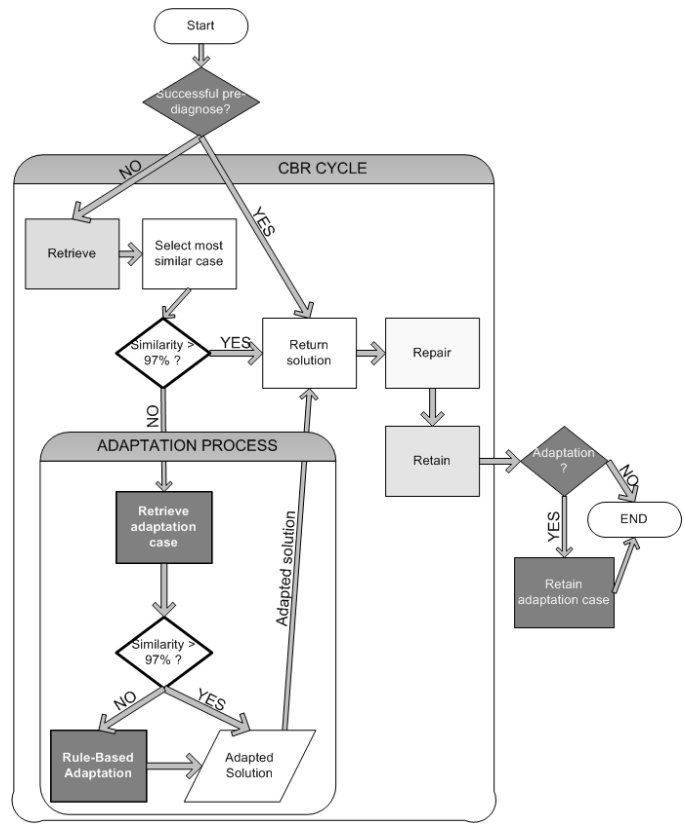

Figure 2. Explanatory image of the developed system

This system receives as input the symptoms shown by the patient. This framework is initially applied to the pre-diagnosis stage. This stage uses a set of basic diagnostic rules so as to identify situations in which a certain set of symptoms indicate the presence of a disease without a shadow of a doubt.

The aim of this stage is to increase the system's performance – it by-passes the complete inference process. Moreover, this is a key stage when the knowledge base is incipient – extremely small – and there is no representative evidence of any of the differential diagnoses in it.

The pre-diagnosis rules have the following structure,

**IF *CSF presents Cloudy aspect* THEN *ABM***

in which it is stated that, if the patient presents "cloudy aspect in the cerebrospinal fluid", then the **unquestionable** existence of Acute Bacterial Meningitis can be affirmed.

The pre-diagnostic rules, as well as the adaptation rules presented later, were developed based on a knowledge engineering work carried out with a physician experienced in the diagnosis of these kinds of diseases.

If the pre-diagnosis stage is successful, then there is a solution to the problem which is presented to the user, offering the possibility of *repairing* the new case.

Once repaired, successful or not, the possibility of *retaining* the solved problematic situation as a new case in the system is offered. In this way is implemented the learning process of the diagnostic experience.

If the case is not obvious or simple, the pre-diagnosis is not applicable and the system proceeds to process the case using the CBR method.

A new query is built from the symptoms of the new case, and the most similar cases are retrieved from the case base – the system retrieves the three most similar cases, and the user can select the one she or he believes more adequate to be reutilized-.

The case representation used consists of three components: a description of the problematic situation, the solution to such situation, and an indicator of success of the solution. The description is comprised of logical attributes (true or false), that indicate the presence or absence of each of the symptoms considered for the diagnosis. The solution is represented by a group of three diagnoses: a primary one and two differential ones.

The retrieval is implemented by the nearest neighbor method using the 81 considered symptoms. The local similarity between symptoms is determined by equality.

After the retrieval, the system evaluates how similar the current case and the selected one are; if the similarity degree exceeds a certain threshold, then the solution to the retrieved case can be reused directly. Otherwise, an adaptation process that integrates CBR with RBR has to be applied.

This process consists of a set of rules that allow carrying out transformations to the solutions. Also, using CBR it is possible to store and reuse change experiences. The process receives as input the differences between the present symptoms in the retrieved case and the current one, as well as the solution offered by the recovered case. From this input, this process offers an adapted solution to the new problem. This solution is the same as the one of the retrieved case, to which some changes were made according to the differences between the problematic situations.

Firstly, a CBR process is carried out to obtain the desired solution, reusing previous change knowledge. For this purpose adaptation cases are used. These consist of two components: description of the case and its solution.

As shown in Figure 3, the difference between the descriptions of the problematic situation, applied to an *S1* solution produces an *S2* solution. This solution is the product of adapting *S1* according to difference *ΔP*.

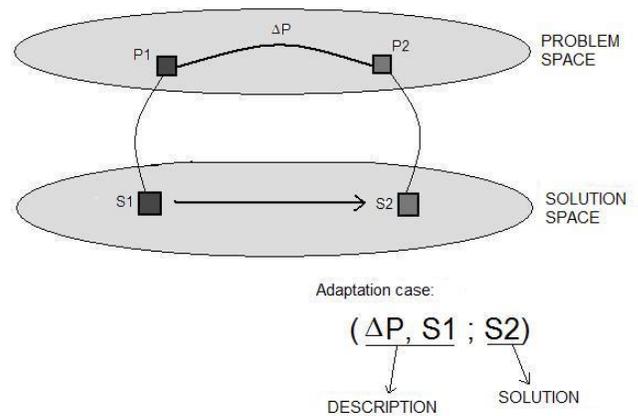

Figure 3.   Adaptation cases structure

The adaptation case description consists of two parts:

a) a set of attributes that represent the differences between the problem descriptions (ΔP). There is one attribute for each considered symptom, which will indicate the following situations: 1) equality between the occurrence of symptoms – with respect to the presence or absence of the symptom in the current and the retrieved case (present or absent in both cases) - ; 2) presence of the symptom in the current case and absence in the previous; and finally, 3) absence of the symptom in the current case and presence in the retrieved one.

b) the solution to the retrieved case, consisting of three possible diagnoses (S1).

The solution (**S2**) to the adaptation case consists of the three diagnoses to which the system arrived once made the pertinent changes to the original solution.

Following a CBR standard process, the first step in the adaptation process consists in the retrieval of the most similar adaptation case. For this purpose is used the nearest neighbor method, with equality functions for the calculation of local similarity. This similarity function only considers those symptoms that, through the process of adaptation knowledge survey carried out with the physicians, were considered as more influential or important to the adaptation process. The symptoms that were not considered for this process are not relevant enough to have an impact in the diagnosis.

Once the most similar adaptation case has been selected, the associated similarity degree is assessed, once again, based on a set threshold. If this threshold is exceeded, the reutilization of the adaptation case is made directly – a null adaptation is done – the adapted solution coincides with the solution of the retrieved case. This is due to the fact that the application context of the retrieved experience and the current context are virtually identical. If the retrieved adaptation case is not similar enough, the system proceeds to adapt based in rules contained in the knowledge base. As well as in the case of the pre-diagnosis stage rules, the domain problem was investigated with the collaboration of an expert doctor. Two main categories of situations have been determined:

- a) Situations in which the differential presence of a symptom allows excluding a certain differential diagnosis. For instance,

IF *Koch's Bacillus* THEN *Discard ABM*

This rule indicates that if the current case (**P2**) shows "Koch's Bacillus", but the retrieved case (**P1**) did not, the differential diagnosis of Acute Bacterial Meningitis is discarded. If such diagnosis was given in the solution of the previous case (**S1**) then it can be removed, and it will not be contained in the solution of the current case. (**S2**)**.**

- b) Situations in which, under the intuition of a given diagnosis and given a difference in the present symptoms another diagnosis is sensed; or its certainty level is modified. For example,

IF ABM *primary* AND *Crystalline aspect* THEN
ABM *Differential*

This rule indicates that if the solution to the previous case (**S1**) presents ABM as a primary diagnosis, but the cases differ in that the current (**P2**) presents "Crystalline aspect in the cerebrospinal fluid" and the previous case (**P1**) does not, then in the current solution (**S2**) such disease has to be considered as a differential diagnosis, not primary.

The different rules modify the facts in the working memory of the inference engine in such way that at the end of the processing the set of resulting diagnoses is retained in it.

The successive application of rules has a correspondence to each of the transformations or adaptations carried out on the original solution, when there is a difference between symptoms.

Once the solution has been obtained (either by the reutilization of the diagnostic case or by its adaptation), it is presented to the user and the system requires her judgment or assessment – success or failure – as for the correction of the suggested diagnosis. If the case fails, the system offers the user the possibility of repairing it, setting the appropriate diagnosis (primary and differentials) – the case is then assumed as successful-.

The next step is the **retention** stage, in which the user is asked if she or he wishes to retain the new case. Moreover, if an adaptation process has been performed, then the user is given the option of storing the adaptation knowledge, which will be available after the new experience. A new adaptation case can be learned either from the reutilization of a previous adaptation case or from the successive application of rules. This type of adaptation process allows, at the beginning, the adaptation to be mostly based in the set of surveyed rules. However, when the system gains experience the reusability of adaptation cases increases, and the system turns less dependent on the rule set.

If the system makes any mistakes when adapting, the user has the possibility of indicating such situation and specifying the correct solution. In this way, the system is able to learn a new adaptation case from the user himself. If this user is a physician with vast experience in this field, this feature is key to the learning of new experiences.

## VI. GENERATION OF THE DIAGNOSIS CASE BASE

One of the main points in the construction of the system consists in obtaining a set of diagnoses large enough – to test the performance of the system – relevant, and representative of real world population.

In previous works of the research group (development of a diagnostic expert system of Acute Bacterial Meningitis based on a Bayesian inference engine [7]), a database of occurrence probabilities of the disease under study (ABM) and its differential diagnoses, based on 10000 medical visits, was available. Likewise, for each disease the corresponding probability of presenting the symptoms under study is available.

Based on these diseases and symptoms probabilities and using the Montecarlo method (assuming normal and uniform distributions in each case) several simulations were performed to obtain a set of virtual "cases".

This initial set was subject to several steps of revision: in first place, the cases with clearly abnormal or extremely improbable situations were removed; and then each of the cases was validated by medical experts. The result is an initial "cured" database that is representative of the real population – it is based on probabilities extracted from the population, and the combinations obtained have been validated by field experts.

Besides the validation process, the expert provided the real diagnosis for each of the cases in the base, for which it can be considered as a set of real cases.

## VII. ASSESSMENT OF THE SYSTEM'S PERFORMANCE

The system's performance has been assessed considering three indicators: accuracy, robustness before partial information, and learning capacity.

### A. Accuracy

In the context of this research, *accuracy* has been defined as the *amount of successful hits* the system reached. For this experiment, a 30-case sample extraction was made (size of the sample calculated according to the population size, based on traditional statistic methods of experiment design). Each case was then presented to the system to obtain the corresponding diagnosis. This diagnosis was later compared to the one provided by the expert to determine whether the system hit or failed. This experiment shown an accuracy of 97%.

### B. Robustness in presence of partial information

This metric intends to assess the system's flexibility when faced to the heterogeneity of the expert users. The potential users of systems of this kind (physicians, students, other specialists in the field of human health) use to have different levels of experience, which implies different capacities in the detection of the symptoms involved. Some symptoms are more difficult to identify than others, and this identification is often related with the experience of the physician. That is to say, in some cases it may happen that the medical user, with short experience in this specific field, might not be able to detect the presence of a certain symptom, even when it is present. The robustness indicator intends to assess how the system behaves when faced to different levels of experience.

The experiment consists in performing a series of cycles, measuring in each one the accuracy of the system, - in the way described above -, but removing from the knowledge base some symptoms in each cycle – symptoms that have been identified as those whose detection is more dependent on the experience of the physician-. The removal of symptoms implies that a certain symptom that used to be considered present is now considered absent, which would coincide with the case entry done by a less experienced doctor.

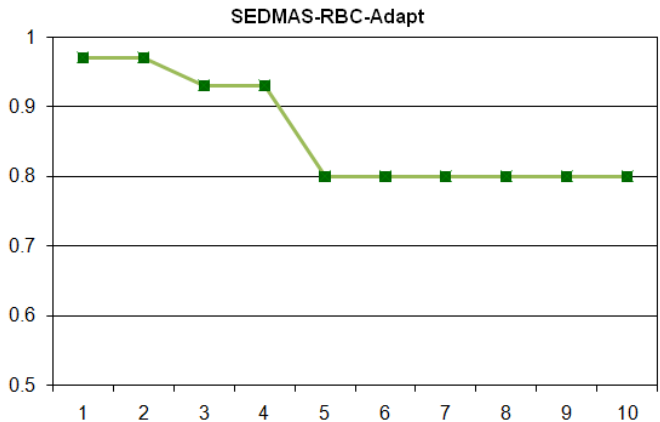

Figure 4.   Results obtained in the experiment of robustness before partial information

In Figure 4 we can see how the accuracy degrades as we remove symptoms. However, it is important to notice that, even though the accuracy of the system decreases, it keeps a constant accuracy of 80% after the fifth iteration, which implies an interesting and desirable result. This behavior has been compared to previous versions developed in the research group using other techniques of intelligent systems (Bayesian inference, CBR without adaptation and without RBR), having obtained better results. [6], [7]

### C. Learning capacity

This metric indicates the system's capacity of incorporating knowledge. *SEDMAS-RBC-Adapt* has important learning capacities. As any CBR system, there is the possibility of retaining new diagnosis cases in the retention stage. Besides, this system has the additional ability to learn new adaptation cases, retaining change experiences.

### VIII.   CONCLUSIONS

The integration of the CBR and RBR methods in the development of a medical diagnosis CDSS has proven to be appropriate and convenient. The approach developed not only presents excellent results for its precision, robustness before partial information and learning capacity, but it is also an example of how to take full advantage of each of the used techniques.

Through the survey of a set of basic rules of pre-diagnosis, the simplest cases to diagnose are detected. Such survey is fairly simple as it is knowledge that doctors handle constantly. For less simple cases in which the diagnosis is not so direct, the system allows resorting to past experience.

In this way, we can take the best out of CBR and RBR.

Also, the approach of the adaptation process is paramount for the system to learn to adapt throughout the time, being able to gather knowledge indirectly from the medical expert. Such approach allows reutilizing the system's adaptation knowledge, while maintaining the capacity to adapt in exceptional cases.